\documentclass[10pt,twocolumn,letterpaper]{article}

\usepackage{cvpr}              

\usepackage{graphicx}
\usepackage{amsmath}
\usepackage{amssymb}
\usepackage{booktabs}
\usepackage{bm}
\usepackage{enumitem}
\usepackage[dvipsnames,table,xcdraw]{xcolor}
\usepackage{wrapfig}

\usepackage{algorithm}
\usepackage[noend]{algpseudocode}

\newcommand{\txt}[1]{{\texttt{#1}}}

\usepackage[T1]{fontenc}
\usepackage{xspace}
\definecolor{green_graph}{RGB}{68, 138, 65}
\definecolor{blue_graph}{RGB}{104, 56, 154}
\definecolor{iODBlue}{RGB}{255, 255, 237}
\definecolor{Gray}{gray}{0.95}
\definecolor{highlight}{RGB}{230, 255, 225}

\newcommand{\ours}{{\fontfamily{cmr}\selectfont ELI}\xspace}

\newcommand{\customsubsubsection}[1]{%
  \par
  \pagebreak[2]%
  \refstepcounter{subsubsection}%
    \everypar={%
      {\setbox0=\lastbox}
      \addcontentsline{toc}{subsubsection}{%
        {\protect\makebox[0.3in][r]{\thesubsubsection.} \hspace*{3pt}#1}}%
      \textbf{\thesubsubsection\space\space{#1}\space}%
      \everypar={}%
    }%
  \ignorespaces
}

\newcommand{\customsubsection}[1]{%
  \par
  \pagebreak[2]%
  \refstepcounter{subsection}%
    \everypar={%
      {\setbox0=\lastbox}
      \addcontentsline{toc}{subsection}{%
        {\protect\makebox[0.3in][r]{\thesubsubsection.} \hspace*{3pt}#1}}%
      \textbf{\thesubsection\space\space{#1}\space}%
      \everypar={}%
    }%
  \ignorespaces
}

\usepackage[pagebackref,breaklinks,colorlinks]{hyperref}

\usepackage[capitalize]{cleveref}
\crefname{section}{Sec.}{Secs.}
\Crefname{section}{Section}{Sections}
\Crefname{table}{Table}{Tables}
\crefname{table}{Tab.}{Tabs.}

\def\cvprPaperID{8375} 
\def\confName{CVPR}
\def\confYear{2022}

\begin{document}

\title{\vspace{-10pt}Energy-based Latent Aligner for Incremental Learning\vspace{-15pt}}


\vspace{-5pt}\author{K J Joseph$^{\dagger \ddag}$~~~~  Salman Khan$^{\ddag \star}$~~~~ Fahad Shahbaz Khan$^{\ddag \diamond}$~~~~ Rao Muhammad Anwer$^{\ddag \P}$ \\ Vineeth N Balasubramanian$^{\dagger}$ \vspace{3pt} \\ 
\normalsize$^{\dagger}$Indian Institute of Technology Hyderabad, India \quad $^{\ddag}$Mohamed bin Zayed University of AI, UAE\\
\normalsize $^{\star}$Australian National University, Australia \quad $^{\diamond}$Linköping University, Sweden \quad $^{\P}$Aalto University, Finland\\
{\tt\small \{cs17m18p100001,~vineethnb\}@iith.ac.in, \{salman.khan,~fahad.khan,~rao.anwer\}@mbzuai.ac.ae}
\vspace{-7pt}
}
\maketitle

\begin{abstract}\vspace{-0.5em}
   Deep learning models tend to forget their earlier knowledge 
   while incrementally learning new tasks.
   This behavior emerges because the parameter updates optimized for the new tasks may not align well with the updates suitable for older tasks. The resulting latent representation mismatch causes forgetting. 
   In this work, we propose \ours: \textbf{E}nergy-based \textbf{L}atent Aligner for \textbf{I}ncremental Learning, which first learns an energy manifold for the latent representations such that previous task latents will have low energy and the current task latents have high energy values. 
   This learned manifold is used to counter the representational shift that happens during incremental learning. 
   The implicit regularization that is offered by our proposed methodology can be used as a 
   plug-and-play module in existing incremental learning methodologies. 
   We validate this through extensive 
   evaluation on CIFAR-100, ImageNet subset, ImageNet 1k and Pascal VOC datasets.
   We observe consistent improvement when \ours is added to three prominent methodologies in class-incremental learning, across multiple incremental settings. Further, when added to the state-of-the-art incremental object detector, \ours provides over $5\%$ improvement in detection accuracy, corroborating its effectiveness and complementary advantage to the existing art. Code is available at: \url{https://github.com/JosephKJ/ELI}.
\end{abstract}

\vspace{-1em}
\section{Introduction}\vspace{-0.5em}
\label{sec:introduction}
Learning experiences are dynamic in the real-world, requiring models to incrementally learn new capabilities over time.  
Incremental Learning (also called continual learning) is a paradigm that learns a model $\mathcal{M}^{\mathcal{T}_{t}}$ at time step $t$, such that it is competent in solving a continuum of tasks $\mathcal{T}_t=\{\tau_1, \tau_2,\cdots,\tau_t\}$ introduced to it during its lifetime. Each task $\tau_i$ contains instances from a disjoint set of classes. Importantly, the training data for the previous tasks $\{\tau_1,\cdots, \tau_{t-1}\}$ cannot be accessed while learning $\tau_{t}$, due to privacy, memory and/or computational constraints. 

\begin{figure}
  \centering
  \includegraphics[width=1\columnwidth]{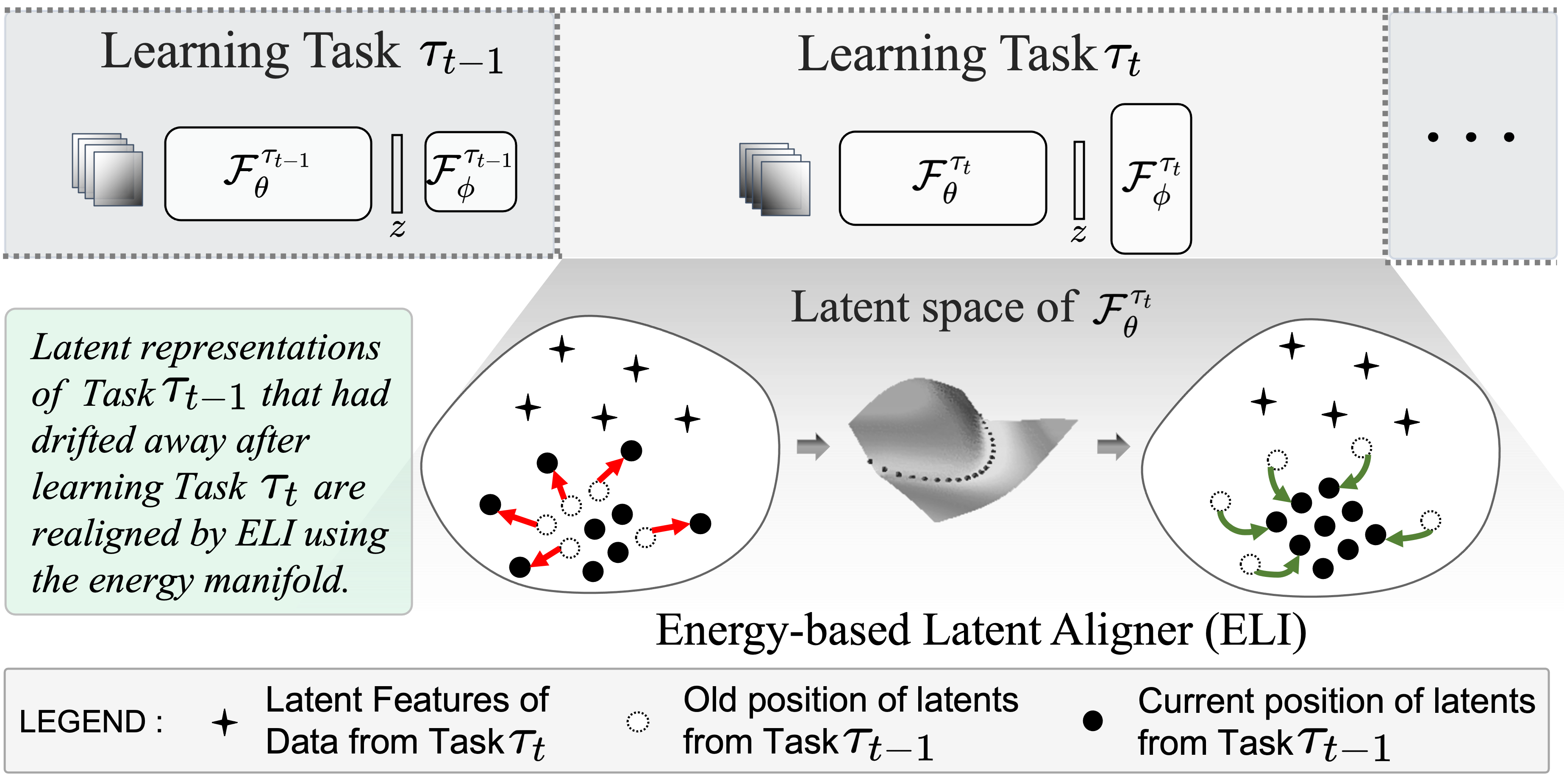}
  \caption{We illustrate an Incremental Learning model trained on a continuum of tasks in the top part of the figure. While learning the current task $\tau_{t}$ (zoomed-in), the latent representation of Task $\tau_{t-1}$ data gets disturbed, as shown by {\color{red}red} arrows. \ours learns an energy manifold, and uses it to counteract this inherent representational shift, as illustrated by {\color{OliveGreen}\textbf{green}} arrows, thereby alleviating forgetting.
  }
  \label{fig:overview}
\end{figure}


We can represent an incremental model $\mathcal{M}^{\mathcal{T}_t}$, as a composition of a latent feature extractor $\mathcal{F}_{\bm \theta}^{\mathcal{T}_t}$ and a trailing network $\mathcal{F}_{\bm \phi}^{\mathcal{T}_t}$ that solves the task using the extracted features: $\mathcal{M}^{\mathcal{T}_t}(\mathbf{x})= (\mathcal{F}_{\bm \phi}^{\mathcal{T}_t} \circ \mathcal{F}_{\bm \theta}^{\mathcal{T}_t})(\mathbf{x})$; where $\mathbf{x} \in \mathcal{T}_{t}$.
A naive approach for learning incrementally would be to use data samples from the current task $\tau_{t}$ to finetune the model trained until the previous task $\mathcal{M}^{\mathcal{T}_{t-1}}$.  Doing so will bias the internal representations of the network to perform well on $\tau_{t}$, in-turn significantly degrading the performance on old tasks. This phenomenon is called catastrophic forgetting \cite{french1999catastrophic,mccloskey1989catastrophic}. 

The incremental learning problem requires accumulating knowledge over a long range of learning tasks without catastrophic forgetting. The main challenge is how to consolidate conflicting implicit representations across different training episodes 
to learn a generalized model applicable to all the learning experiences.  To this end, existing approaches investigate
regularization-based methods \cite{aljundi2018memory,kirkpatrick2017overcoming,li2018learning,zenke2017continual,kj2020meta,rajasegaran2020itaml} that  constrain ${\bm \theta}$ and ${\bm \phi}$ such that the model performs well on all the tasks. Exemplar replay-based methods \cite{AGEM,rebuffi2017icarl,lopez2017gradient,castro2018end,joseph2021towards} retain a subset of datapoints from each task, and rehearse them to learn a continual model. Dynamically expanding models \cite{mallya2018packnet,serra2018overcoming,rusu2016progressive}, enlarge ${\bm \theta}$ and ${\bm \phi}$ while learning incrementally.

Complementary to the existing methodologies, we introduce a novel approach which minimizes the representational shift in the latent space of an incremental model, using a learned energy manifold. 
The energy modeling offers a natural mechanism to deal with catastrophic forgetting which we build upon.
Fig.~\ref{fig:overview} illustrates how our proposed methodology, \ours: \textbf{E}nergy-based \textbf{L}atent Aligner for \textbf{I}ncremental Learning, helps to alleviate forgetting. 
After learning the current task $\tau_t$, the features from the feature extractor (referred to as \textit{latents} henceforth), of the previous task data
$\mathbf{z}^{\mathcal{T}_{t}} = \mathcal{F}_{{\bm \theta}}^{\mathcal{T}_{t}}(\mathbf{x})$, $\mathbf{x} \in \tau_{t-1}$ drift as shown by the {\color{red}{red}} arrows. The \emph{first} step in our approach is to learn an energy manifold where the latent representations from the model trained until the current task $\mathcal{M}^{\mathcal{T}_t}$ have higher energy, while the latents from the model trained till the previous task $\mathcal{M}^{\mathcal{T}_{t-1}}$ have lower energy. \emph{Next}, the learned energy-based model (EBM) is used to transform the previous task latents $\mathbf{z}^{\mathcal{T}_{t}}$ (obtained via passing the previous task data through current model) which had drifted away, to alternate locations in the latent space such that the representational shift is undone (as shown by the {\color{OliveGreen}\textbf{green}} arrows). This helps alleviate forgetting in incremental learning. We explain how this transformation can be achieved in Sec.~\ref{sec:methodology}. We also present a proof-of-concept with MNIST (Fig.~\ref{fig:proof-of-concept}) which mimics the above setting. 
The latent space visualization and accuracy regain after learning the new task 
correlates with the illustration in Fig.~\ref{fig:overview}, which reinforces our intuition.

A unique characteristic of our energy-based latent aligner is its ability to extend and enhance existing continual learning methodologies, without any change to their methodology. We verify this by adding \ours to three prominent class-incremental methods: iCaRL \cite{rebuffi2017icarl}, LUCIR \cite{hou2019learning} and AANet \cite{liu2021adaptive} and the state-of-the-art incremental Object Detector: iOD \cite{iOD}. We conduct thorough experimental evaluation on incremental versions of large-scale classification datasets like CIFAR-100 \cite{krizhevsky2009cifar}, ImageNet subset \cite{rebuffi2017icarl} and  ImageNet 1k \cite{deng2009imagenet}; and Pascal VOC \cite{everingham2010pascal} object detection dataset. For incremental classification experiments, we consider two prominent setups: adding classes to a model trained with half of all the classes as first task, and the general incremental learning setting which considers equal number of classes for all tasks. 
\ours consistently improves 
performance across all datasets and on all methods in incremental classification settings, and obtains impressive performance gains on incremental Object Detection,
compared to current state-of-the-art \cite{iOD}, by  $5.4\%$, $7\%$ and $3\%$ while incrementally learning $10$, $5$ and a single class respectively.

\noindent To summarize, the key highlights of our work are:

\begin{itemize}[leftmargin=*,topsep=0pt, noitemsep]
\item We introduce a novel methodology \ours, which helps to counter the representational shift that happens in the latent space of incremental learning models.
\item Our energy-based latent aligner can act as an add-on module to existing incremental classifiers and object detectors, without any changes to their methodology.
\item \ours shows consistent improvement on over $45$ experiments across three large scale incremental classification datasets, and improves the current state-of-the-art incremental object detector by over $5\%$ mAP
on average.\end{itemize}


\section{Related Work}\vspace{-0.5em}
\label{sec:relatedworks}

\noindent\textbf{Incremental Learning:} In this setting a model consistently improves itself on new tasks, without compromising its performance on old tasks.  
One popular approach to achieve this behaviour is by {constraining} the parameters to not deviate much from previously tuned values \cite{li2017learning,rebuffi2017icarl,castro2018end,wu2019large,douillard2020podnet,liu2020mnemonics}. 
In this regard, knowledge distillation \cite{hinton2015distilling} has been used extensively to enforce explicit regularization in incremental classification \cite{li2017learning,rebuffi2017icarl,castro2018end} and object detection \cite{shmelkov2017incremental,joseph2021towards,gupta2021ow} settings.
In replay based methods, typically a small subset of exemplars is stored to recall and retain representations useful for earlier tasks \cite{rebuffi2017icarl,liu2020mnemonics,belouadah2019il2m,hou2019learning,kj2020meta}. Another set of {isolated parameter} learning methods dedicate separate subsets of parameters to different tasks, thus avoiding interference \eg, by new network blocks or gating mechanisms
\cite{rusu2016progressive,rajasegaran2019adaptive,rajasegaran2019random,abati2020conditional,liu2021adaptive}. Further, meta-learning approaches have been explored to learn the update directions which are shared among multiple incremental tasks \cite{riemer2018learning,rajasegaran2020itaml,iOD}. In contrast to these approaches, we propose to learn an EBM to align implicit feature distributions between incremental tasks. \ours can enhance these existing methods without any methodological modifications, by enforcing an implicit latent space regularization using the learned energy manifold.

\noindent\textbf{Energy-based Models:} EBMs \cite{lecun2006tutorial} are a type of maximum likelihood estimation models that can assign low energies to observed data-label pairs and high energies otherwise \cite{NEURIPS2019_378a063b}. 
 EBMs have been used for out-of-distribution sample detection \cite{liu2020energy,9533706}, structured prediction \cite{belanger2016structured,tu-18,JMLR:v22:20-326} and improving adversarial robustness \cite{hill2021stochastic,NEURIPS2019_378a063b}. Joint Energy-based Model (JEM) \cite{grathwohl2019your} shows that any classifier can be reinterpreted as a generative model that can model the joint likelihood of labels and data. While JEM requires alternating between a discriminative and generative objective, Wang \etal \cite{wang2021energy} propose an energy-based open-world softmax objective that can jointly perform discriminative learning and generative modeling. 
 EBMs have also been used for synthesizing images \cite{xiao2021vaebm,arbel2021generalized,zhao2020learning,zhao2021unpaired}. 
 Xie \etal \cite{xie2016theory} represents EBM using a CNN and utilizes Langevin dynamics for MCMC
 sampling to generate realistic images. 
 In contrast to these methods, we explore the utility of the EBMs to alleviate forgetting in a continual learning paradigm. 
 Most of these methods operate in the data space, where sampling from the EBM would be expensive\cite{zhao2021unpaired}. Differently, we learn the energy manifold with the latent representations, which is faster and effective in controlling the representational shift that affects incremental models. 
 A recent unpublished work \cite{li2020energy} proposes to replace the standard softmax layer of an incremental model with an energy-based classifier head. Our approach introduces an implicit regularization in the latent space using the learned energy manifold which is fundamentally different from their approach, scales well to harder datasets and diverse settings (classification and detection).

\section{Energy-based Latent Aligner}
\label{sec:methodology}

Our proposed methodology \ours utilizes an Energy-based Model (EBM) \cite{lecun2006tutorial} to optimally adapt the latent representations of an incremental model, such that it alleviates catastrophic forgetting. We refer to the intermediate feature vector extracted from the backbone network of the model as \emph{latent} representations in our discussion. After a brief introduction to the problem setting in Sec.~\ref{sec:problem_setting}, we explain how the EBM is learned and used for aligning in Sec.~\ref{sec:latent_aligner}. We conclude with a discussion on a toy experiment in Sec.~\ref{sec:toy_example}.

\subsection{Problem Setting}\label{sec:problem_setting}
In the incremental learning paradigm, a set of tasks $\mathcal{T}_t=\{\tau_1, \tau_2,\cdots,\tau_t\}$ is introduced to the model over time.
$\tau_t$ denotes the task introduced at time step $t$, which is composed of images $\mathbf{X}^{\tau_t}$ and labels $\mathbf{y}^{\tau_t}$ sampled from its corresponding task data distribution: $(\mathbf{x}^{\tau_{t}}_i, y^{\tau_{t}}_i) \sim p_{data}^{\tau_t}$.
Each task $\tau_t$, contains instances from a disjoint set of classes.
We seek to build a model $\mathcal{M}^{\mathcal{T}_t}$, which is competent in solving all the tasks $\mathcal{T}_t$. 
Without loss of generality $\mathcal{M}^{\mathcal{T}_t}$ can be expressed as a composition of two functions: $\mathcal{M}^{\mathcal{T}_t}(\mathbf{x})= (\mathcal{F}_{\bm \phi}^{\mathcal{T}_t} \circ \mathcal{F}_{\bm \theta}^{\mathcal{T}_t})(\mathbf{x})$, where $\mathcal{F}_{\bm \theta}^{\mathcal{T}_t}$ is a feature extractor and  $\mathcal{F}_{\bm \phi}^{\mathcal{T}_t}$ is a classifier in the case of a classification model and a composite  classification and localization branch for an object detector, solving all the tasks $\mathcal{T}_t$ introduced to it so far. 

While training $\mathcal{M}^{\mathcal{T}_t}$ on current task $\tau_t$, the model does not have access to all the data from previous tasks\footnote{Such restricted memory is considered due to practical limitations such as bounded storage, computational budget and privacy issues.}. This imbalance between present and previous task data can bias the model to focus on the latest task, while catastrophically degrading its performance on the earlier ones. Making an incremental learner robust against such forgetting is a challenging research question. 
Regularization methods\cite{aljundi2018memory,kirkpatrick2017overcoming}, exemplar-replay methods \cite{AGEM,rebuffi2017icarl,lopez2017gradient} and progressive model expansion methods\cite{mallya2018packnet,serra2018overcoming,rusu2016progressive} have emerged as the standard ways to address forgetting. Our proposed methodology is complementary to all these developments in the field, and is generic enough to serve as an add-on to any such continual learning methodology, with minimal overhead. 

\subsection{Latent Aligner}\label{sec:latent_aligner}
We perform energy-based modeling in the latent space of continual learning models. 
Our 
latent aligner approach avoids the need to explicitly  identify which latent representations should be adapted or retained to preserve
knowledge across tasks while learning new skills. 
It implicitly identifies which representations are ideal to be shared between 
tasks, preserves them, and simultaneously adapts representations which negatively impact incremental learning.


Let us consider a concrete incremental learning setting where we introduce a new task $\tau_{t}$ to a model that is trained to perform well until the previous tasks $\mathcal{M}^{\mathcal{T}_{t-1}}$. Training data to learn the new task is sampled from the corresponding data distribution: $(\mathbf{x}^{\tau_{t}}_i, y^{\tau_{t}}_i) \sim p_{data}^{\tau_t}$. We may use any existing continual learning algorithm $\mathcal{A}$, to learn an incremental model $\mathcal{M}^{\mathcal{T}_t}$. The latent representations of $\mathcal{M}^{\mathcal{T}_{t-1}}$ would be 
optimized for learning $\tau_{t}$, which causes degraded performance of $\mathcal{M}^{\mathcal{T}_{t}}$ on the previous tasks. Depending on the efficacy of $\mathcal{A}$, $\mathcal{M}^{\mathcal{T}_{t}}$ can have varying degrees of effectiveness in alleviating the inherent forgetting. 
Our proposed method helps to undo this representational shift that happens to previous task instances, when passed through $\mathcal{M}^{\mathcal{T}_{t}}$.

\begin{figure}
  \centering
  \includegraphics[width=1\columnwidth]{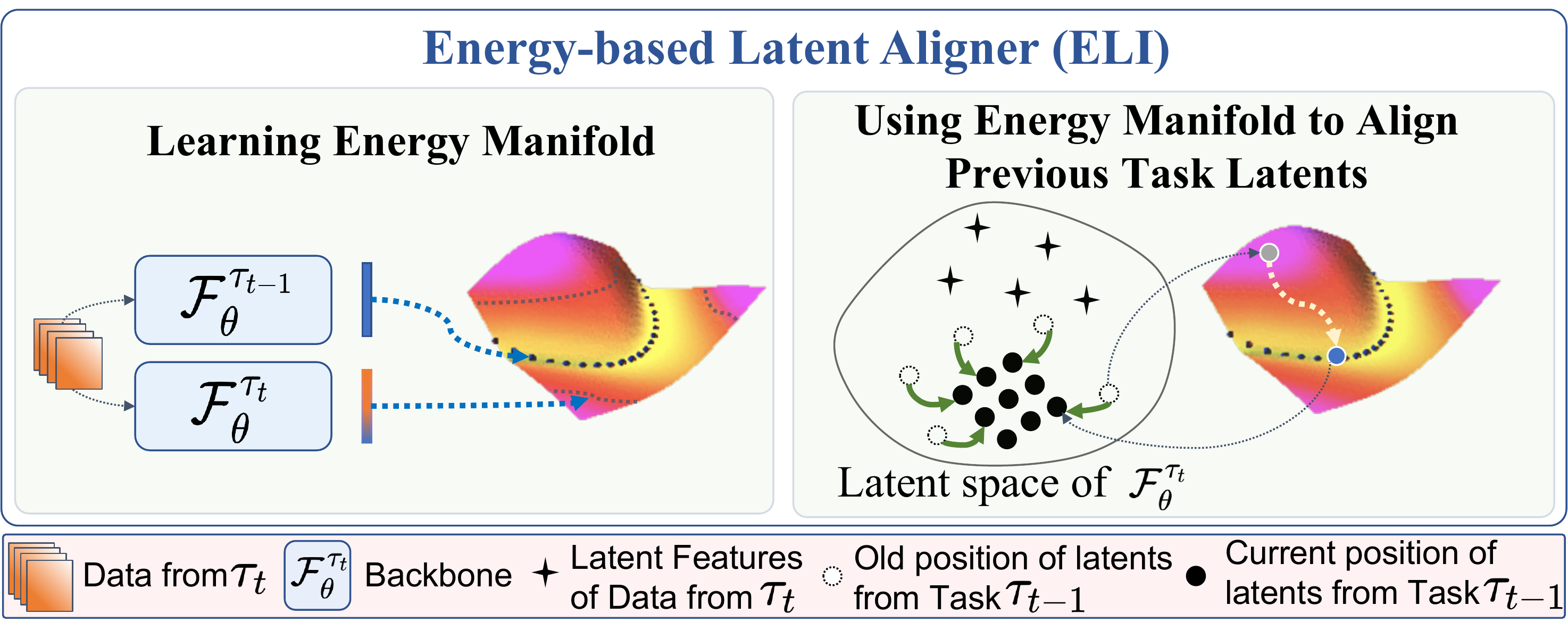}
  \caption{
  We learn an energy manifold using the latent representations of the current task data passed though the current model $\mathcal{F}_{{\bm \theta}}^{\mathcal{T}_{t}}$ and previous model $\mathcal{F}_{{\bm \theta}}^{\mathcal{T}_{t-1}}$. This manifold is used to align the latents from $\tau_{t-1}$ that were shifted while learning the new task.}
  \label{fig:learning_and_using_manifold}
\end{figure}

As illustrated in Fig.~\ref{fig:learning_and_using_manifold}, in the \emph{first} step, we learn an energy manifold using three ingredients: (i) images from the current task: $\mathbf{x} \sim p_{data}^{\tau_{t}}$, (ii) latent representations of $\mathbf{x}$ from the model trained till previous task: $\mathbf{z}^{\mathcal{T}_{t-1}} = \mathcal{F}_{\bm \theta}^{\mathcal{T}_{t-1}}(\mathbf{x})$ and (iii) latent representations of $\mathbf{x}$ from the model trained till the current task: $\mathbf{z}^{\mathcal{T}_{t}} = \mathcal{F}_{\bm \theta}^{\mathcal{T}_{t}}(\mathbf{x})$. 
An energy-based model $E_{\psi}$ is learned to assign low energy values for $\mathbf{z}^{\mathcal{T}_{t-1}}$, and high energy values for $\mathbf{z}^{\mathcal{T}_{t}}$.
\emph{Next}, during inference, the learned energy manifold $E_{\psi}$ is used to counteract the representational shift that happens to the latent representations of previous task instances when passed through the current model: $\mathbf{z}^{\mathcal{T}_{t}} = \mathcal{F}_{{\bm \theta}}^{\mathcal{T}_{t}}(\mathbf{x})$ where $\mathbf{x} \in \mathcal{T}_{t-1}$. 
Due to the representational shift in the latent space, $\mathbf{z}^{\mathcal{T}_{t}}$ will have higher energy values in the energy manifold. We align $\mathbf{z}^{\mathcal{T}_{t}}$ to alternate locations in latent space such that their energy on the manifold is minimized, as illustrated in right part of Fig.~\ref{fig:learning_and_using_manifold}. These shifted latents demonstrate less forgetting, which we empirically verify through large scale experiments on incremental classification and object detection in Sec.~\ref{sec:experiments}. 

It is interesting to note the following: 1) Our method adds implicit regularization in the latent space without making any changes to the incremental learning algorithm $\mathcal{A}$, which is used to learn $\mathcal{M}^{\mathcal{T}_{t}}$, 2) \ours does not require access to previous task data to learn the energy manifold. Current task data, passed though the model 
$\mathcal{F}_{{\bm \theta}}^{\mathcal{T}_{t-1}}$ indeed acts as a proxy for previous task data while learning the EBM.


\vspace{0.5em}
\customsubsubsection{Learning the Latent Aligner:~}\label{sec:learning_RAIL}
EBMs provide a simple and flexible way to model data likelihoods \cite{NEURIPS2019_378a063b}. We use continuous energy-based models, formulated using a neural network, which can generically model a diverse range of function mappings. 
Specifically, for a given latent feature vector $\mathbf{z} \in \mathbb{R}^D$ in \ours, we learn an energy function $E_\psi(\mathbf{z}):\mathbb{R}^D \rightarrow \mathbb{R}$ to map it to a scalar energy value.
An EBM is defined as Gibbs distribution $p_\psi(\mathbf{z})$ over $E_\psi(\mathbf{z})$:
\begin{equation}
    p_{\psi}(\mathbf{z}) = \frac{\text{exp~}(-E_\psi(\mathbf{z}))}{\int_{\mathbf{z}} \text{exp~}(-E_\psi(\mathbf{z})) d\mathbf{z}},  \label{eqn:ebm}
\end{equation}
where $\int_{\mathbf{z}} \text{exp~}(-E_\psi(\mathbf{z})) d\mathbf{z}$ is an intractable partition function. 
EBM is trained by maximizing the data log-likelihood on a sample set drawn from the true 
distribution $p_{true}(\mathbf{z})$:
\begin{equation}
    L(\psi) = \mathbb{E}_{\mathbf{z}\sim p_{true}}[\log p_{\psi}(\mathbf{z})].
\end{equation}
The derivative of the above objective is as follows \cite{woodford2006notes}:
\begin{equation}
    \partial_\psi L(\psi) = \mathbb{E}_{\mathbf{z}\sim p_{true}}[-\partial_\psi E_\psi(\mathbf{z})] + \mathbb{E}_{\mathbf{z}\sim p_{\psi}}[\partial_\psi E_\psi(\mathbf{z})].
    \label{eqn:gradient}
\end{equation}

The first term in Eq.~\ref{eqn:gradient} ensures that the energy for a sample $\mathbf{z}$ drawn from the true data distribution $p_{true}$ will be minimized, while the second term ensures that the samples drawn from the model itself, will have higher energy values. In \ours, $p_{true}$ corresponds to the distribution of latent representations from the model trained till the previous task at any point in time.  Sampling from $p_\psi(\mathbf{x})$ is intractable owing to the normalization constant in Eq.~\ref{eqn:ebm}. Approximate samples are recursively drawn using Langevin dynamics \cite{neal2011mcmc, welling2011bayesian}, which is a popular MCMC algorithm, 
\begin{equation}
    \mathbf{z}_{i+1} = \mathbf{z}_{i} - \frac{\lambda}{2}\partial_{\mathbf{z}} E_{\psi}(\mathbf{z}) + \sqrt{\lambda}\omega_i \text{, } \omega_i \sim \mathcal{N}(0, \mathbf{I})
    \label{eqn:sampling}
\end{equation}
where $\lambda$ is the step size and $\omega$ captures data uncertainty. Eq.~\ref{eqn:sampling} yields a Markov chain that stabilizes to an stationary distribution within few 
iterations, starting from an initial 
$\mathbf{z}_i$.

Algorithm \ref{algo:learning_ebm} illustrates how the energy manifold is learned in \ours.
The energy function $E_{\psi}$ is realised by a multi-layer perceptron with a single neuron in the output layer, which quantifies energy of the input sample. It is Kaiming initialized in Line 1.
Until a few number of iterations, we sample mini-batches from the current task data distribution $p_{data}^{\tau_t}$. Next, the latent representation of the data in the mini-batch is retrieved from the model trained until the previous task $\mathcal{F}_{{\bm \theta}}^{\mathcal{T}_{t-1}}$ and the model trained till the current task $\mathcal{F}_{{\bm \theta}}^{\mathcal{T}_{t}}$, in Line 4 and 5 respectively. 
From here on, we prepare to compute the gradients according to Eq.~\ref{eqn:gradient}, which is required for training the energy function.
The first term in Eq.~\ref{eqn:gradient} minimizes the expectation over in-distribution energies, which is computed in Line 7, while the second term maximizes the expectation over out-of-distribution energies (Line 8). The Langevin sampling which is required to compute the out-of-distribution energies takes the latents from the current model as initial starting points of the Markov chain, as illustrated in Line 6. Finally, the loss is computed in Line 9 and the energy function $E_\psi$ is optimized with RMSprop~\cite{hinton2012neural} optimizer in Line 10.

\begin{algorithm}
\caption{Algorithm \textsc{LearnEBM}}
\label{algo:learning_ebm}
\begin{algorithmic}[1]
\Require{Feature extractor of model trained till current task: $\mathcal{F}_{\bm \theta}^{\mathcal{T}_t}$; Feature extractor of model trained till previous task: $\mathcal{F}_{\bm \theta}^{\mathcal{T}_{t-1}}$; Data distribution of the current task: $p_{data}^{\tau_t}$
}
\State $E_\psi \leftarrow $Initialize the Energy function. 
\While{until required iterations}
\State $\mathbf{x} \sim p_{data}^{\tau_t}$ \Comment{\textit{Sample a mini-batch}}
\State $\mathbf{z}^{\mathcal{T}_{t-1}} \leftarrow \mathcal{F}_{{\bm \theta}}^{\mathcal{T}_{t-1}} (\mathbf{x})$
\State $\mathbf{z}^{\mathcal{T}_{t}} \leftarrow \mathcal{F}_{{\bm \theta}}^{\mathcal{T}_{t}} (\mathbf{x})$
\State $\mathbf{z}^{\mathcal{T}_{t}}_{sampled} \leftarrow$ Sample from EBM with $\mathbf{z}^{\mathcal{T}_{t}}$ as starting $~~~~~~~~~~~~~~~~~~~~~~~~~~~$points. \Comment{\textit{Refer Equation~\ref{eqn:sampling}}}
\State in\_dist\_energy $\leftarrow E_{\psi}(\mathbf{z}^{\mathcal{T}_{t-1}})$
\State out\_of\_dist\_energy $\leftarrow E_{\psi}(\mathbf{z}^{\mathcal{T}_{t}}_{sampled})$
\State $Loss$ $\leftarrow $ ($-$in\_dist\_energy $+$ out\_of\_dist\_energy) 
$~~~~~~~~~$ \Comment{\textit{Refer Equation~\ref{eqn:gradient}}}
\State Optimize $E_\psi$ with $Loss$.
\EndWhile
\State \Return $E_\psi$
\end{algorithmic}
\end{algorithm}

\customsubsubsection{Alignment using \ours:~}\label{sec:aligning_with_RAIL}
After learning a task $\tau_t$ in an incremental setting, we use Algorithm \ref{algo:learning_ebm} to learn the energy manifold. This manifold is used to align the latent representations of previous task instances 
from the current model $\mathcal{M}^{\mathcal{T}_{t}}$ using Algorithm \ref{algo:align_latents}. 
The gradient of energy function $E_{\psi}$ with respect to the latent representation $\mathbf{z}$ is computed (Line 2). These latents are then successively updated to reduce their energy (Line 3). We repeat this for $L_{steps}$ number of Langevin iterations.
The aligner assumes that a high-level task information is available during inference i.e., whether a latent belongs to the current task or not.


\begin{algorithm}
\caption{Algorithm \textsc{AlignLatents}}
\label{algo:align_latents}
\begin{algorithmic}[1]
\Require{Latent vector to be adapted: $\mathbf{z}$; EBM: $E_{\psi}$; Number of Langevin steps: $L_{steps}$; Learning rate: $\lambda$
}
\While{until $L_{steps}$ iterations}
\State $grad \leftarrow \nabla_{\mathbf{z}} E_{\psi}(\mathbf{z})$
\State $\mathbf{z} \leftarrow \mathbf{z} - \lambda * grad$
\EndWhile
\State \Return $\mathbf{z}$
\end{algorithmic}
\end{algorithm}

\begin{figure*}
\centering
\includegraphics[width=2.12\columnwidth]{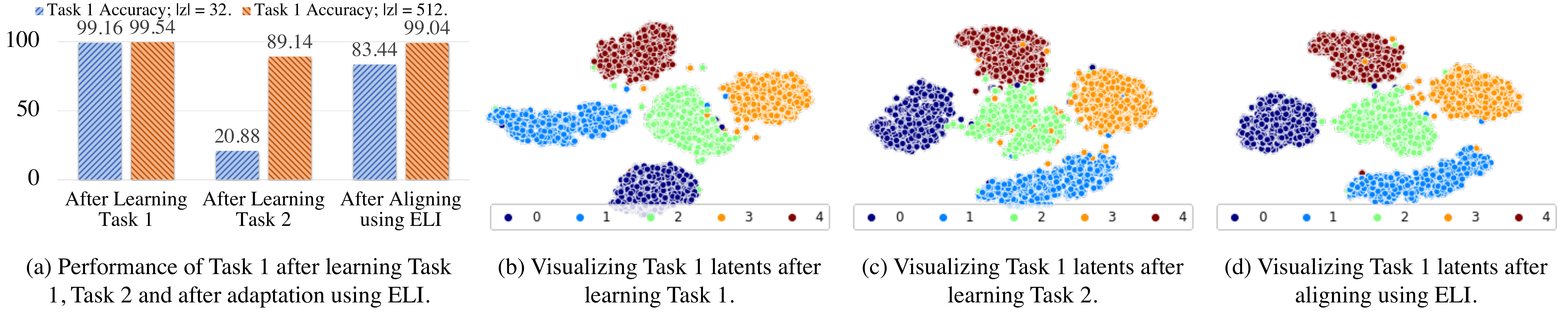}
\caption{A key hypothesis that we base our methodology is that while learning a new task, the latent representations will get disturbed, which will in-turn cause catastrophic forgetting of the previous task, and that an energy manifold can be used to align these latents, such that it alleviates forgetting. Here, we illustrate a proof-of-concept that our hypothesis is indeed true.
We consider a two task experiment on MNIST; $\mathcal{T}_1 = \{\tau_0,\tau_1,\tau_2,\tau_3,\tau_4\}, \mathcal{T}_2 = \{\tau_5,\tau_6,\tau_7,\tau_8,\tau_9\}$. After learning the second task, the accuracy on $\mathcal{T}_1$ test set drops to $20.88\%$, while experimenting with a $32$-dim latent space.
The latent aligner in \ours provides $62.56\%$ improvement in test accuracy to $83.44\%$.
The visualization of a $512$-dim latent space after learning $\mathcal{T}_2$ in sub-figure (c), indeed shows cluttering due to representational shift. \ours is able to align the latents as shown in sub-figure (d), which alleviates the drop in accuracy from $89.14\%$ to $99.04\%$.
}
\label{fig:proof-of-concept}
\end{figure*}

\vspace{-10pt}
\subsection{Toy Example}\label{sec:toy_example}
Our  methodology is build on a key premise that latent representations of an incremental learning model will get disturbed after training on new tasks, and that an energy-based manifold can aid in successfully mitigating this unwarranted representational shift in a post-hoc fashion. In Fig.~\ref{fig:proof-of-concept}, we present a proof-of-concept that our hypothesis indeed holds. 
We consider a two task experiment with incremental MNIST, where the first task is to learn the first 5 classes, while the second is to learn the rest: $\mathcal{T}_1 = \{\tau_0 \ldots \tau_4\}$ and $\mathcal{T}_2 = \{\tau_5 \ldots\tau_9\}$. 
We first learn $\mathcal{M}^{\mathcal{T}_1}(\mathbf{x})= (\mathcal{F}_{\bm \phi}^{\mathcal{T}_1} \circ \mathcal{F}_{\bm \theta}^{\mathcal{T}_1})(\mathbf{x})$, where $\mathbf{x} \in \mathcal{T}_1$, and then incrementally update it to $\mathcal{M}^{\mathcal{T}_2}(\mathbf{x})= (\mathcal{F}_{\bm \phi}^{\mathcal{T}_2} \circ \mathcal{F}_{\bm \theta}^{\mathcal{T}_2})(\mathbf{x})$, where $\mathbf{x} \in \mathcal{T}_2$. 
When evaluating the Task 1 classification accuracy using $(\mathcal{F}_{\bm \phi}^{\mathcal{T}_1} \circ \mathcal{F}_{\bm \theta}^{\mathcal{T}_2})(\mathbf{x})$, where $\mathbf{x} \in \mathcal{T}_1^{test}$, we see catastrophic forgetting in action. There is a significant drop in performance from $99.2\%$ to $20.9\%$, when we use a $32$ dimensional latent space.
Let $\mathcal{F}_{\psi}^{\text{\ours}}$ represent our proposed latent aligner. 
While re-evaluating the classification accuracy using $(\mathcal{F}_{\bm \phi}^{\mathcal{T}_1} \circ \mathcal{F}_{\psi}^{\text{\ours}} \circ \mathcal{F}_{\bm \theta}^{\mathcal{T}_2})(\mathbf{x})$, where $\mathbf{x} \in \mathcal{T}_1^{test}$, we see an improvement of $62.6\%$ to $83.4\%$.
We also try increasing the latent space dimension to $512$. 
Consistent to our earlier observation, we observe a drop in accuracy from $99.54\%$ to $89.14\%$. \ours helps to improve it to $99.04\%$. The absolute drop in performance due to forgetting is lower than the $32$ dimensional latent space because of the larger capacity of the model. 
The visualization of latent space in sub-figure (c) also suggests more cluttering.
Sub-figure (d) explicitly reinforces the utility of \ours to realign the latents. Specifically, note how \txt{Class 3} latents which were intermingled with \txt{Class 2} latents are now nicely moved around in the latent space by \ours.
These results strongly motivate the utility of our method. 
By making $\mathcal{F}_{\bm \theta}^{\mathcal{T}_2}$ more stronger using mainstream incremental learning methodologies, we would improve the performance further. We illustrate this on harder datasets for class-incremental learning and incremental object detection setting in Sec.~\ref{sec:CIL_results} and Sec.~\ref{sec:iOD_results} respectively.

\section{Experiments and Results}
\label{sec:experiments}
We conduct extensive experiments with incremental {classifiers} and object {detectors} to evaluate 
\ours.
To the best of our knowledge, ours is the first methodology, which works across both these settings \emph{without} any modification. 

\noindent \textbf{Protocols:}
In both problem domains, we study class-incremental setting where a group of classes constitutes an incremental task.
For class-incremental learning of classifiers, we experiment with two prominent protocols that exist in the literature: \textbf{a)} train with half the total number of classes as the first task \cite{liu2021adaptive,hou2019learning}, and equal number of classes per task thereafter, \textbf{b)} ensure that each task (including the first) has equal number of classes \cite{rebuffi2017icarl, castro2018end, rajasegaran2019random, kj2020meta}. The former tests extreme class incremental learning setting, where in the $25$ task setting we incrementally add only two classes at each stage for a dataset with $100$ classes. It has the advantage of learning a strong initial classifier as it has access to half of the dataset in Task 1. The later setting has a uniform class distribution across tasks. Both these settings test different plausible dynamics of an incremental classifier. For incremental object detection, similar to existing works \cite{iOD,shmelkov2017incremental,peng2020faster}, we follow a two task setting where the second task contains $10$, $5$ or a single incremental class.

\begin{table*}[]
\centering\setlength{\tabcolsep}{2pt}
\caption{The table shows class-incremental learning results when our latent aligner \ours is added to three prominent and top-performing incremental approaches \cite{rebuffi2017icarl,hou2019learning,liu2021adaptive}. 
\ours is able to provide additional latent space regularization to these methods, consistently improving them across all the settings. The {\color{ForestGreen}{green}} subscript highlights the relative improvement. Refer to Sec.~\ref{sec:CIL_results} for detailed analysis.
}
\label{tab:cil_results}
\resizebox{\textwidth}{!}{%
\begin{tabular}{>{\kern-\tabcolsep}ll|llllll|llllll<{\kern-\tabcolsep}}
\toprule
\rowcolor{Gray}
\multicolumn{2}{c|}{Settings $\rightarrow$} & \multicolumn{6}{c|}{Half of all the classes is used to learn the first task} & \multicolumn{6}{c}{Same number of classes for each task} \\ \midrule
\multicolumn{2}{c|}{Datasets $\rightarrow$} & \multicolumn{3}{c|}{CIFAR-100} & \multicolumn{3}{c|}{ImageNet subset} & \multicolumn{3}{c|}{CIFAR-100} & \multicolumn{3}{c}{ImageNet subset} \\ \midrule
\multicolumn{1}{c}{Methods} & \multicolumn{1}{c|}{Venue} & \multicolumn{1}{c}{5 Tasks} & \multicolumn{1}{c}{10 Tasks} & \multicolumn{1}{c|}{25 Tasks} & \multicolumn{1}{c}{5 Tasks} & \multicolumn{1}{c}{10 Tasks} & \multicolumn{1}{c|}{25 Tasks} & \multicolumn{1}{c}{5 Tasks} & \multicolumn{1}{c}{10 Tasks} & \multicolumn{1}{c|}{20 Tasks} & \multicolumn{1}{c}{5 Tasks} & \multicolumn{1}{c}{10 Tasks} & \multicolumn{1}{c}{20 Tasks} \\ \midrule
\multicolumn{1}{l}{iCaRL \cite{rebuffi2017icarl}} & CVPR 17 & $56.97$ & $53.28$ & \multicolumn{1}{l|}{$50.98$} & $58.24$ & $51.6$ & $49.02$ & $61.59$ & $60.05$ & \multicolumn{1}{l|}{$57.81$} & $71.46$ & $65.25$ & $60.21$ \\
\multicolumn{1}{l}{iCaRL + \ours} &  & $63.68_{\color{ForestGreen}\textbf{~+~6.71}}$ & $58.92_{\color{ForestGreen}\textbf{~+~5.64}}$ & \multicolumn{1}{l|}{$54.00_{\color{ForestGreen}\textbf{~+~3.02}}$} & $68.94_{\color{ForestGreen}\textbf{~+~10.73}}$ & $61.48_{\color{ForestGreen}\textbf{~+~9.88}}$ & $56.11_{\color{ForestGreen}\textbf{~+~7.08}}$ & $\textbf{70.13}_{\color{ForestGreen}\textbf{~+~8.54}}$ & $\textbf{67.81}_{\color{ForestGreen}\textbf{~+~7.75}}$ & \multicolumn{1}{l|}{$\textbf{63.06}_{\color{ForestGreen}\textbf{~+~5.25}}$} & $\textbf{78.51}_{\color{ForestGreen}\textbf{~+~7.04}}$ & $\textbf{71.66}_{\color{ForestGreen}\textbf{~+~6.41}}$ & $\textbf{66.77}_{\color{ForestGreen}\textbf{~+~6.56}}$ \\
\multicolumn{1}{l}{} &  &  &  & \multicolumn{1}{l|}{} &  &  &  &  &  & \multicolumn{1}{l|}{} &  &  &  \\
\multicolumn{1}{l}{LUCIR \cite{hou2019learning}} & CVPR 19 & $64.37$ & $62.57$ & \multicolumn{1}{l|}{$59.91$} & $71.38$ & $68.99$ & $64.65$ & $62.01$ & $58.95$ & \multicolumn{1}{l|}{$54.2$} & $74.22$ & $67.97$ & $62.2$ \\
\multicolumn{1}{l}{LUCIR + \ours} &  & $66.06_{\color{ForestGreen}\textbf{~+~1.69}}$ & $63.50_{\color{ForestGreen}\textbf{~+~0.93}}$ & \multicolumn{1}{l|}{$60.30_{\color{ForestGreen}\textbf{~+~0.39}}$} & $\textbf{74.58}_{\color{ForestGreen}\textbf{~+~3.21}}$ & $71.62_{\color{ForestGreen}\textbf{~+~2.61}}$ & $66.35_{\color{ForestGreen}\textbf{~+~1.71}}$ & $64.55_{\color{ForestGreen}\textbf{~+~2.49}}$ & $59.51_{\color{ForestGreen}\textbf{~+~0.56}}$ & \multicolumn{1}{l|}{$54.98_{\color{ForestGreen}\textbf{~+~0.78}}$} & $75.38_{\color{ForestGreen}\textbf{~+~1.16}}$ & $70.28_{\color{ForestGreen}\textbf{~+~2.31}}$ & $65.51_{\color{ForestGreen}\textbf{~+~3.31}}$ \\
\multicolumn{1}{l}{} &  &  &  & \multicolumn{1}{l|}{} &  &  &  &  &  & \multicolumn{1}{l|}{} &  &  &  \\
\multicolumn{1}{l}{AANet \cite{liu2021adaptive}} & CVPR 21 & $67.53$ & $66.25$ & \multicolumn{1}{l|}{$64.28$} & $70.84$ & $70.3$ & $69.07$ & $63.89$ & $60.94$ & \multicolumn{1}{l|}{$56.88$} & $65.86$ & $54.13$ & $44.96$ \\
\multicolumn{1}{l}{AANet + \ours} &  & $\textbf{68.78}_{\color{ForestGreen}\textbf{~+~1.25}}$ & $\textbf{66.62}_{\color{ForestGreen}\textbf{~+~0.37}}$ & \multicolumn{1}{l|}{$\textbf{64.72}_{\color{ForestGreen}\textbf{~+~0.44}}$} & $73.54_{\color{ForestGreen}\textbf{~+~2.73}}$ & $\textbf{71.82}_{\color{ForestGreen}\textbf{~+~1.52}}$ & $\textbf{70.32}_{\color{ForestGreen}\textbf{~+~1.25}}$ & $66.36_{\color{ForestGreen}\textbf{~+~2.47}}$ & $61.72_{\color{ForestGreen}\textbf{~+~0.78}}$ & \multicolumn{1}{l|}{$57.65_{\color{ForestGreen}\textbf{~+~0.77}}$} & $67.43_{\color{ForestGreen}\textbf{~+~1.57}}$ & $55.47_{\color{ForestGreen}\textbf{~+~1.34}}$ & $46.93_{\color{ForestGreen}\textbf{~+~1.97}}$ \\ \bottomrule
\end{tabular}%
}
\end{table*}

\begin{figure*}
\includegraphics[width=2.1\columnwidth]{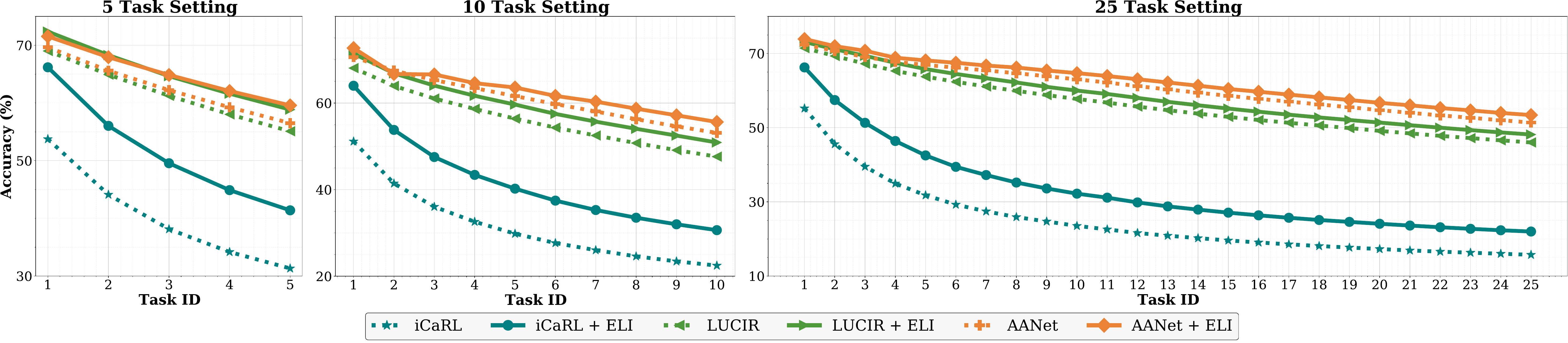}
\caption{ Here, we plot the average accuracy after learning each incremental task on ImageNet 1k dataset. \ours is able to consistently improve iCaRL~\cite{rebuffi2017icarl}, LUCIR~\cite{hou2019learning} and AANet~\cite{liu2021adaptive} on $5$ task, $10$ task and $25$ task setting. On average, we see $8.17\%$, $3.05\%$ and $2.53\%$ improvement to the three base methods. \emph{(Best viewed in color)}
}
\label{fig:imagenet}
\end{figure*}

\begin{table*}[]
\centering
\caption{Incremental Object Detection is evaluated in a two task setting with Pascal VOC 2007 dataset \cite{everingham2010pascal}. We consider adding $10$, $5$ and one class (highlighted in {\colorbox{ iODBlue}{color})} to a detector trained on the rest of the classes. When added to the state-of-the-art incremental Object Detector iOD~\cite{iOD}, \ours provide a competitive improvement of $5.4 \%$, $7 \%$ and $3 \%$ mAP in $10+10$, $15+5$ and $19+1$ settings respectively.
}
\label{tab:iOD_results}
\resizebox{\textwidth}{!}{%
\begin{tabular}{>{\kern-\tabcolsep}llllllllllllllllllllll<{\kern-\tabcolsep}}
\toprule
\rowcolor{Gray} ${\color{ForestGreen}\textbf{10 + 10 Setting}}$ & aero & cycle & bird & boat & bottle & bus & car & cat & chair & cow & table & dog & horse & bike & person & plant & sheep & sofa & train & tv & mAP \\ \midrule
All 20 & 79.4 & 83.3 & 73.2 & 59.4 & 62.6 & 81.7 & 86.6 & 83 & 56.4 & 81.6 & \cellcolor[HTML]{ffffed}71.9 & \cellcolor[HTML]{ffffed}83 & \cellcolor[HTML]{ffffed}85.4 & \cellcolor[HTML]{ffffed}81.5 & \cellcolor[HTML]{ffffed}82.7 & \cellcolor[HTML]{ffffed}49.4 & \cellcolor[HTML]{ffffed}74.4 & \cellcolor[HTML]{ffffed}75.1 & \cellcolor[HTML]{ffffed}79.6 & \cellcolor[HTML]{ffffed}73.6 & 75.2 \\
First 10 & 78.6 & 78.6 & 72 & 54.5 & 63.9 & 81.5 & 87 & 78.2 & 55.3 & 84.4 & \cellcolor[HTML]{ffffed}- & \cellcolor[HTML]{ffffed}- & \cellcolor[HTML]{ffffed}- & \cellcolor[HTML]{ffffed}- & \cellcolor[HTML]{ffffed}- & \cellcolor[HTML]{ffffed}- & \cellcolor[HTML]{ffffed}- & \cellcolor[HTML]{ffffed}- & \cellcolor[HTML]{ffffed}- & \cellcolor[HTML]{ffffed}- & 73.4 \\
Std Training & 35.7 & 9.1 & 16.6 & 7.3 & 9.1 & 18.2 & 9.1 & 26.4 & 9.1 & 6.1 & \cellcolor[HTML]{ffffed}57.6 & \cellcolor[HTML]{ffffed}57.1 & \cellcolor[HTML]{ffffed}72.6 & \cellcolor[HTML]{ffffed}67.5 & \cellcolor[HTML]{ffffed}73.9 & \cellcolor[HTML]{ffffed}33.5 & \cellcolor[HTML]{ffffed}53.4 & \cellcolor[HTML]{ffffed}61.1 & \cellcolor[HTML]{ffffed}66.5 & \cellcolor[HTML]{ffffed}57 & 37.3 \\ \midrule
Shmelkov \etal \cite{shmelkov2017incremental} & 69.9 & 70.4 & 69.4 & 54.3 & 48 & 68.7 & 78.9 & 68.4 & 45.5 & 58.1 & \cellcolor[HTML]{ffffed}59.7 & \cellcolor[HTML]{ffffed}72.7 & \cellcolor[HTML]{ffffed}73.5 & \cellcolor[HTML]{ffffed}73.2 & \cellcolor[HTML]{ffffed}66.3 & \cellcolor[HTML]{ffffed}29.5 & \cellcolor[HTML]{ffffed}63.4 & \cellcolor[HTML]{ffffed}61.6 & \cellcolor[HTML]{ffffed}69.3 & \cellcolor[HTML]{ffffed}62.2 & 63.1 \\
Faster ILOD \cite{peng2020faster} & 72.8 & 75.7 & 71.2 & 60.5 & 61.7 & 70.4 & 83.3 & 76.6 & 53.1 & 72.3 & \cellcolor[HTML]{ffffed}36.7 & \cellcolor[HTML]{ffffed}70.9 & \cellcolor[HTML]{ffffed}66.8 & \cellcolor[HTML]{ffffed}67.6 & \cellcolor[HTML]{ffffed}66.1 & \cellcolor[HTML]{ffffed}24.7 & \cellcolor[HTML]{ffffed}63.1 & \cellcolor[HTML]{ffffed}48.1 & \cellcolor[HTML]{ffffed}57.1 & \cellcolor[HTML]{ffffed}43.6 & 62.2 \\
ORE~\cite{joseph2021towards} & 63.5 & 70.9 & 58.9 & 42.9 & 34.1 & 76.2 & 80.7 & 76.3 & 34.1 & 66.1 & \cellcolor[HTML]{ffffed}56.1 & \cellcolor[HTML]{ffffed}70.4 & \cellcolor[HTML]{ffffed}80.2 & \cellcolor[HTML]{ffffed}72.3 & \cellcolor[HTML]{ffffed}81.8 & \cellcolor[HTML]{ffffed}42.7 & \cellcolor[HTML]{ffffed}71.6 & \cellcolor[HTML]{ffffed}68.1 & \cellcolor[HTML]{ffffed}77 & \cellcolor[HTML]{ffffed}67.7 & 64.6 \\ \midrule
iOD \cite{iOD} & 76 & 74.6 & 67.5 & 55.9 & 57.6 & 75.1 & 85.4 & 77 & 43.7 & 70.8 & \cellcolor[HTML]{ffffed}60.1 & \cellcolor[HTML]{ffffed}66.4 & \cellcolor[HTML]{ffffed}76 & \cellcolor[HTML]{ffffed}72.6 & \cellcolor[HTML]{ffffed}74.6 & \cellcolor[HTML]{ffffed}39.7 & \cellcolor[HTML]{ffffed}64 & \cellcolor[HTML]{ffffed}60.2 & \cellcolor[HTML]{ffffed}68.5 & \cellcolor[HTML]{ffffed}60.5 & 66.3 \\
iOD + \ours & 78.5 & 81.6 & 73.8 & 65.5 & 63.2 & 80.2 & 87.7 & 82.5 & 52.4 & 81.2 & \cellcolor[HTML]{ffffed}55.5 & \cellcolor[HTML]{ffffed}73.1 & \cellcolor[HTML]{ffffed}80.5 & \cellcolor[HTML]{ffffed}76.5 & \cellcolor[HTML]{ffffed}80.4 & \cellcolor[HTML]{ffffed}42.2 & \cellcolor[HTML]{ffffed}68.8 & \cellcolor[HTML]{ffffed}66 & \cellcolor[HTML]{ffffed}72.6 & \cellcolor[HTML]{ffffed}70.8 & \textbf{71.7} \\ \midrule \midrule
${\color{ForestGreen}\textbf{15 + 5 Setting}}$ & aero & cycle & bird & boat & bottle & bus & car & cat & chair & cow & table & dog & horse & bike & person & plant & sheep & sofa & train & tv & mAP \\ \midrule
All 20 & 79.4 & 83.3 & 73.2 & 59.4 & 62.6 & 81.7 & 86.6 & 83 & 56.4 & 81.6 & 71.9 & 83 & 85.4 & 81.5 & 82.7 & \cellcolor[HTML]{ffffed}49.4 & \cellcolor[HTML]{ffffed}74.4 & \cellcolor[HTML]{ffffed}75.1 & \cellcolor[HTML]{ffffed}79.6 & \cellcolor[HTML]{ffffed}73.6 & 75.2 \\
First 15 & 78.1 & 82.6 & 74.2 & 61.8 & 63.9 & 80.4 & 87 & 81.5 & 57.7 & 80.4 & 73.1 & 80.8 & 85.8 & 81.6 & 83.9 & \cellcolor[HTML]{ffffed}- & \cellcolor[HTML]{ffffed}- & \cellcolor[HTML]{ffffed}- & \cellcolor[HTML]{ffffed}- & \cellcolor[HTML]{ffffed}- & 53.2 \\
Std Training & 12.7 & 0.6 & 9.1 & 9.1 & 3 & 0 & 8.5 & 9.1 & 0 & 3 & 9.1 & 0 & 3.3 & 2.3 & 9.1 & \cellcolor[HTML]{ffffed}37.6 & \cellcolor[HTML]{ffffed}51.2 & \cellcolor[HTML]{ffffed}57.8 & \cellcolor[HTML]{ffffed}51.5 & \cellcolor[HTML]{ffffed}59.8 & 16.8 \\ \midrule
Shmelkov \etal \cite{shmelkov2017incremental} & 70.5 & 79.2 & 68.8 & 59.1 & 53.2 & 75.4 & 79.4 & 78.8 & 46.6 & 59.4 & 59 & 75.8 & 71.8 & 78.6 & 69.6 & \cellcolor[HTML]{ffffed}33.7 & \cellcolor[HTML]{ffffed}61.5 & \cellcolor[HTML]{ffffed}63.1 & \cellcolor[HTML]{ffffed}71.7 & \cellcolor[HTML]{ffffed}62.2 & 65.9 \\
Faster ILOD \cite{peng2020faster} & 66.5 & 78.1 & 71.8 & 54.6 & 61.4 & 68.4 & 82.6 & 82.7 & 52.1 & 74.3 & 63.1 & 78.6 & 80.5 & 78.4 & 80.4 & \cellcolor[HTML]{ffffed}36.7 & \cellcolor[HTML]{ffffed}61.7 & \cellcolor[HTML]{ffffed}59.3 & \cellcolor[HTML]{ffffed}67.9 & \cellcolor[HTML]{ffffed}59.1 & 67.9 \\
ORE~\cite{joseph2021towards} & 75.4 & 81 & 67.1 & 51.9 & 55.7 & 77.2 & 85.6 & 81.7 & 46.1 & 76.2 & 55.4 & 76.7 & 86.2 & 78.5 & 82.1 & \cellcolor[HTML]{ffffed}32.8 & \cellcolor[HTML]{ffffed}63.6 & \cellcolor[HTML]{ffffed}54.7 & \cellcolor[HTML]{ffffed}77.7 & \cellcolor[HTML]{ffffed}64.6 & 68.5 \\ \midrule
iOD \cite{iOD} & 78.4 & 79.7 & 66.9 & 54.8 & 56.2 & 77.7 & 84.6 & 79.1 & 47.7 & 75 & 61.8 & 74.7 & 81.6 & 77.5 & 80.2 & \cellcolor[HTML]{ffffed}37.8 & \cellcolor[HTML]{ffffed}58 & \cellcolor[HTML]{ffffed}54.6 & \cellcolor[HTML]{ffffed}73 & \cellcolor[HTML]{ffffed}56.1 & 67.8 \\
iOD + \ours & 80.1 & 85.8 & 73.6 & 68.8 & 66.3 & 85.2 & 87.5 & 84.1 & 59.9 & 81.2 & 74.6 & 83.7 & 85.3 & 77.9 & 80.3 & \cellcolor[HTML]{ffffed}45.2 & \cellcolor[HTML]{ffffed}63.4 & \cellcolor[HTML]{ffffed}66.2 & \cellcolor[HTML]{ffffed}77.6 & \cellcolor[HTML]{ffffed}69.5 & \textbf{74.8} \\ \midrule \midrule
${\color{ForestGreen}\textbf{19 + 1 Setting}}$ & aero & cycle & bird & boat & bottle & bus & car & cat & chair & cow & table & dog & horse & bike & person & plant & sheep & sofa & train & tv & mAP \\ \midrule
All 20 & 79.4 & 83.3 & 73.2 & 59.4 & 62.6 & 81.7 & 86.6 & 83 & 56.4 & 81.6 & 71.9 & 83 & 85.4 & 81.5 & 82.7 & 49.4 & 74.4 & 75.1 & 79.6 & \cellcolor[HTML]{ffffed}73.6 & 75.2 \\
First 19 & 76.3 & 77.3 & 68.4 & 55.4 & 59.7 & 81.4 & 85.3 & 80.3 & 47.8 & 78.1 & 65.7 & 77.5 & 83.5 & 76.2 & 77.2 & 46.6 & 71.4 & 65.8 & 76.5 & \cellcolor[HTML]{ffffed}- & 67.5 \\
Std Training & 16.6 & 9.1 & 9.1 & 9.1 & 9.1 & 8.3 & 35.3 & 9.1 & 0 & 22.3 & 9.1 & 9.1 & 9.1 & 13.7 & 9.1 & 9.1 & 23.1 & 9.1 & 15.4 & \cellcolor[HTML]{ffffed}50.7 & 14.3 \\ \midrule
Shmelkov \etal \cite{shmelkov2017incremental} & 69.4 & 79.3 & 69.5 & 57.4 & 45.4 & 78.4 & 79.1 & 80.5 & 45.7 & 76.3 & 64.8 & 77.2 & 80.8 & 77.5 & 70.1 & 42.3 & 67.5 & 64.4 & 76.7 & \cellcolor[HTML]{ffffed}62.7 & 68.3 \\
Faster ILOD \cite{peng2020faster} & 64.2 & 74.7 & 73.2 & 55.5 & 53.7 & 70.8 & 82.9 & 82.6 & 51.6 & 79.7 & 58.7 & 78.8 & 81.8 & 75.3 & 77.4 & 43.1 & 73.8 & 61.7 & 69.8 & \cellcolor[HTML]{ffffed}61.1 & 68.6 \\
ORE~\cite{joseph2021towards} & 67.3 & 76.8 & 60 & 48.4 & 58.8 & 81.1 & 86.5 & 75.8 & 41.5 & 79.6 & 54.6 & 72.8 & 85.9 & 81.7 & 82.4 & 44.8 & 75.8 & 68.2 & 75.7 & \cellcolor[HTML]{ffffed}60.1 & 68.9 \\ \midrule
iOD \cite{iOD} & 78.2 & 77.5 & 69.4 & 55 & 56 & 78.4 & 84.2 & 79.2 & 46.6 & 79 & 63.2 & 78.5 & 82.7 & 79.1 & 79.9 & 44.1 & 73.2 & 66.3 & 76.4 & \cellcolor[HTML]{ffffed}57.6 & 70.2 \\
iOD + \ours & 84.7 & 79.2 & 73.7 & 60.1 & 61.8 & 82.8 & 85.4 & 82.9 & 51.3 & 82.7 & 64.5 & 82.3 & 82.9 & 75.9 & 78.7 & 50.7 & 73.9 & 74.7 & 76.7 & \cellcolor[HTML]{ffffed}59.2 & \textbf{73.2} \\ \bottomrule
\end{tabular}%
}
\end{table*}

\noindent \textbf{Datasets and Evaluation Metrics:}\label{sec:datasets}
Following existing works \cite{liu2021adaptive,hou2019learning,rebuffi2017icarl, castro2018end,iOD,shmelkov2017incremental} we use incremental versions of CIFAR-100 \cite{krizhevsky2009cifar}, ImageNet subset \cite{rebuffi2017icarl},  ImageNet 1k \cite{deng2009imagenet} and Pascal VOC \cite{everingham2010pascal} datasets. CIFAR-100 \cite{krizhevsky2009cifar} contains $50$k training images, corresponding to $100$ classes, each with spatial dimensions of $32\times32$. ImageNet-subset \cite{rebuffi2017icarl} contains $100$ randomly selected classes from ImageNet datasets. We also experiment with the full ImageNet 2012 dataset \cite{deng2009imagenet} which contains $1000$ classes. In contrast to CIFAR-100, there are over $1300$ images per class with $224\times224$ size in both ImageNet-subset and ImageNet-1k.  Pascal VOC 2007 \cite{everingham2010pascal} contains $9963$ images, where each object instance is annotated with its class label and location in an image. Instances from $20$ classes are annotated in Pascal VOC. 
Average accuracy across tasks \cite{rebuffi2017icarl,liu2021adaptive} and mean average precision (mAP) \cite{everingham2010pascal} is used as the evaluation metric for incremental classification and detection, respectively.

\begin{figure*}
\includegraphics[width=2.1\columnwidth]{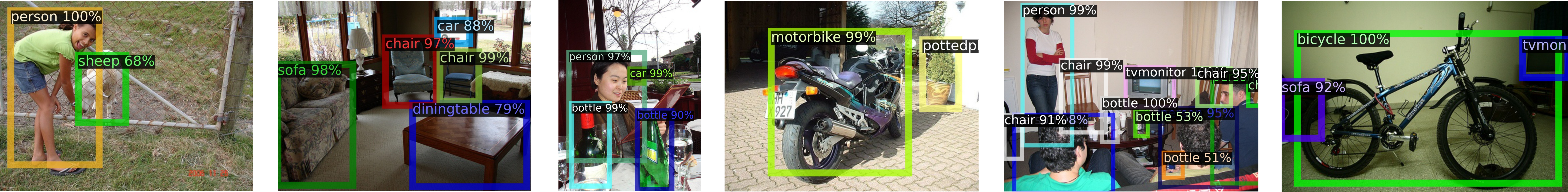}
\caption{In these qualitative results of incremental Object Detection,  instances of  \txt{plant},  \txt{sheep},  \txt{sofa},  \txt{train} and \txt{tvmonitor} were 
introduced to a detector trained on the rest.
We detect instances of old and new classes alike. More results are in supplementary materials.
}
\label{fig:qualitative_results}
\end{figure*}

\noindent \textbf{Implementation Details:}\label{sec:implementation_details}
Following the standard practice \cite{rebuffi2017icarl,liu2021adaptive}, we use ResNet-18 \cite{he2016identity} for CIFAR-100 experiments and ResNet-32 \cite{he2016identity} for ImageNet experiments. We use a batch size of $128$ and train for $160$ epochs. We start with an initial learning rate of $0.1$, which is decayed by $0.1$ after $80^{th}$ and $120^{th}$ epochs. 
The EBM is a three layer neural network with $64$ neurons in the first two layers and single neuron in the last layer. The 
features that are passed on to the final softmax classifier of the base network, are used for learning the EBM.  It is trained for $1500$ iterations with mini-batches of size $128$. The learning rate is set to $0.0001$. We use $30$ langevin iterations to sample from the EBM. We found that keeping an exponential moving average of the EBM model
was effective.
The implementations of the three prominent class-incremental methodologies (iCaRL \cite{rebuffi2017icarl}, LUCIR \cite{hou2019learning} and AANet \cite{liu2021adaptive}) follows the official code from AANet \cite{liu2021adaptive} authors, released under an MIT license. They use an exemplar store of $20$ images per class. Note that our latent aligner does not use exemplars. The iCaRL inference is modified to use fully-connected layers following Castro \etal \cite{castro2018end}. 
All results are mean of three runs.
We use an incremental version of Faster R-CNN \cite{ren2015faster} for object detection experiments, following iOD \cite{iOD}. The $2048$ dimensional penultimate feature vector from the RoI Head is used to learning the EBM. 

\subsection{Incremental Classification Results}\label{sec:CIL_results}
We augment three popular class-incremental learning methods: iCaRL~\cite{rebuffi2017icarl}, LUCIR~\cite{hou2019learning} and AANet~\cite{liu2021adaptive} with our proposed latent aligner. Table \ref{tab:cil_results} showcases the results on CIFAR-100 \cite{krizhevsky2009cifar} and ImageNet subset \cite{rebuffi2017icarl} datasets. As explained earlier, we conduct experiments on the setting where half of the classes are learned in the first task, and when all tasks has equal number of classes. In the former, we group $10$, $5$ and $2$ classes each to create $5$, $10$ and $25$ learning tasks respectively, after training the model on $50$ initial classes. In the second setting, we group $20$, $10$ and $5$ classes each to create $5$, $10$ and $20$ incremental tasks.
We see consistent improvement across all these settings when we add \ours to the corresponding base methodology. In both the settings, the improvement is more pronounced on harder datasets. 
LUCIR~\cite{hou2019learning} and AANet~\cite{liu2021adaptive} use an explicit latent space regularizer in their methodology. \ours is able to improve them further. Simpler methods like iCaRL~\cite{rebuffi2017icarl} benefit more from the implicit regularization that \ours offers (this aspect is explored further in Sec.~\ref{sec:implicit_regularizer}). In Fig.~\ref{fig:imagenet}, we plot the average accuracy after learning each task in $5$ task, $10$ task and $25$ task
settings on ImageNet 1k. We see a similar trend, but with larger improvements on this harder dataset. When added to iCaRL~\cite{rebuffi2017icarl}, LUCIR~\cite{hou2019learning} and AANet~\cite{liu2021adaptive}, \ours provides $8.17\%$, $3.05\%$ and $2.53\%$ improvement on average in ImageNet 1k experiments, respectively. 

When we consider adding same number of classes in each incremental task, simple logit distillation provided by  iCaRL~\cite{rebuffi2017icarl}, along with our proposed latent aligner outperforms complicated methods by a significant margin. This is because the feature learning that happens with half of the classes in the first task, is a major prerequisite for good performance of approaches like LUCIR~\cite{hou2019learning} and AANet~\cite{liu2021adaptive}. 

\subsection{Incremental Object Detection Results}\label{sec:iOD_results}
Following the standard evaluation protocol \cite{iOD,shmelkov2017incremental} for incremental object detection, we group classes from Pascal VOC 2007 \cite{everingham2010pascal} into two tasks. Three different task combinations are considered here. We initially learn $10$, $15$ or $19$ classes, and then introduce $10$, $5$ or one class as the second task,  respectively. Table \ref{tab:iOD_results} shows the results of this experiment. The first two rows in each section give the upper-bound and the accuracy after learning the first task. 
The `Std Training' row shows how the performance on previous classes deteriorate when simply finetuning the model on the new class instances. The next three rows titled Shmelkov \etal \cite{shmelkov2017incremental}, Faster ILOD \cite{peng2020faster} and ORE \cite{joseph2021towards} show how existing methods help to address catastrophic forgetting. We add \ours to iOD \cite{iOD}, the current state-of-the-art method, to improve its mAP by $5.4\%$, $7\%$ and $3\%$ while adding $10$, $5$ and one class respectively, to a detector trained on the rest. This improvement can be attributed to the effectiveness of \ours in aligning the latent representations to reduce forgetting. 
These results also 
demonstrate that \ours is an effective plug-and-play method to reduce forgetting, across classification and detection tasks. Fig.~\ref{fig:qualitative_results} shows our qualitative results.

\begin{table*}
\parbox{.275\linewidth}{
\centering
\caption{We vary the number of Langevin steps $L_{steps}$, required to sample from the EBM. The latents get aligned even within a few steps.}
\label{tab:langevin}
\resizebox{.275\textwidth}{!}{%
\begin{tabular}{>{\kern-\tabcolsep}cccc<{\kern-\tabcolsep}}
\toprule
\rowcolor{Gray}
\# of steps & 5 Tasks & 10 Tasks & 25 Tasks \\ \midrule
5 & 63.30 & 58.61 & 52.81 \\
10 & 63.66 & 58.85 & 53.49 \\
20 & 63.63 & 58.90 & 53.76 \\
\rowcolor{highlight} 
30 & 63.68 & 58.92 & 54.00 \\
60 & 63.73 & 59.01 & 54.07 \\
90 & 63.79 & 58.97 & 54.04 \\ \bottomrule
\end{tabular}%
}
}
\hspace{20pt}
\parbox{.295\linewidth}{
\centering
\caption{We change the number of iterations for training the EBM in Algo.~\ref{algo:learning_ebm}. The EBM converges within $1$k iterations, with moderate improvement thereafter.}
\label{tab:iterations}
\resizebox{.295\textwidth}{!}{%
\begin{tabular}{>{\kern-\tabcolsep}cccc<{\kern-\tabcolsep}}
\toprule
 \rowcolor{Gray}
\# of iterations & 5 Tasks & 10 Tasks & 25 Tasks \\ \midrule
10 & 56.90 & 53.85 & 49.02 \\
100 & 60.53 & 57.08 & 50.41 \\
1000 & 63.60 & 58.88 & 53.66 \\
\rowcolor{highlight} 
1500 & 63.68 & 58.92 & 54.00 \\
2000 & 63.80 & 58.97 & 54.03 \\
3000 & 63.67 & 58.85 & 54.06 \\ \bottomrule
\end{tabular}%
}
}
\hspace{20pt}
\parbox{.32\linewidth}{
\centering
\caption{We vary the architecture of the EBM here. \texttt{i} and \texttt{o} refers to input and output layer, while the values in-between represent the number of neurons in each layer.}
\label{tab:architecture}
\resizebox{.32\textwidth}{!}{%
\begin{tabular}{>{\kern-\tabcolsep}cccc<{\kern-\tabcolsep}}
\toprule
\rowcolor{Gray}
Architecture & 5 Tasks & 10 Tasks & 25 Tasks \\ \midrule
i - o & 60.97 & 57.52 & 53.92 \\
i - 64 - o & 63.72 & 59.02 & 54.59 \\
\rowcolor{highlight} 
i - 64 - 64 - o & 63.68 & 58.92 & 54.00 \\
i - 64 - 64 - 64 - o & 63.71 & 58.9 & 54.44 \\
i - 256 - 256 - o & 63.53 & 58.68 & 54.16 \\
i - 512 - 512 - o & 63.66 & 58.66 & 54.00 \\ \bottomrule
\end{tabular}%
}
}
\vspace{-10pt}
\end{table*}

\section{Discussions and Analysis}
\label{sec:ablations}
\customsubsection{\ours as an Implicit Regularizer:}\label{sec:implicit_regularizer}
\begin{wrapfigure}[12]{r}{0.27\textwidth}
\vspace{-27pt}
  \begin{center}
    \includegraphics[width=0.27\textwidth]{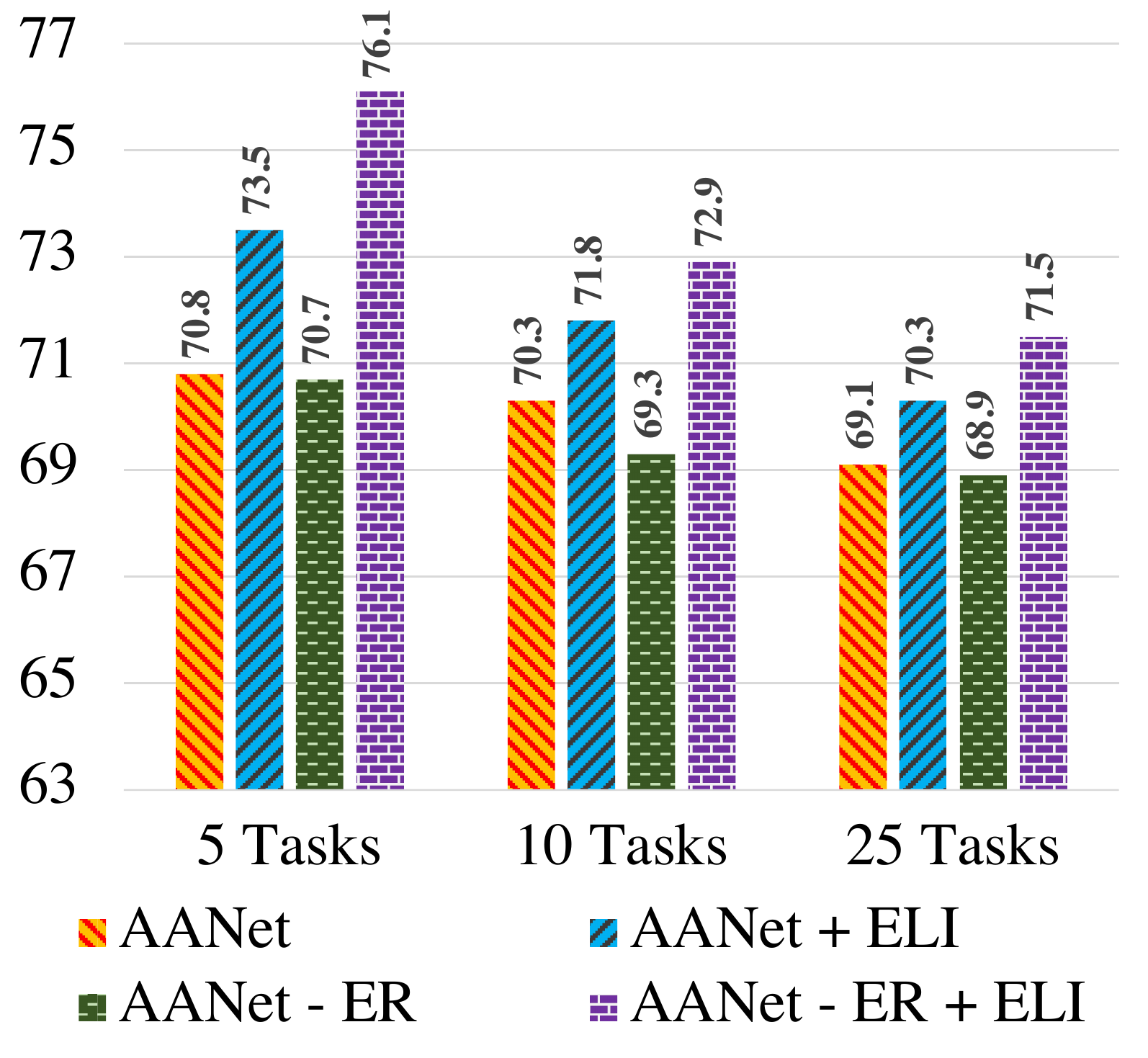}
  \end{center}
  \vspace{-12pt}
  \caption{\ours as an Implicit Regularizer on ImageNet subset.}
  \label{fig:implicit_reg}
\end{wrapfigure}

To showcase the effectiveness of the implicit regularization that \ours offers, we remove the explicit latent regularization term (referred to as ER in Fig.~\ref{fig:implicit_reg}) from our top performing method AANet \cite{liu2021adaptive} on ImageNet subset \cite{rebuffi2017icarl} experiments. There is a consistent drop in accuracy when ER is removed from the base method ({\color{green_graph}\textbf{green}} bars). \ours is able to improve the performance of such a model by $5.41\%$, $3.58\%$ and $2.57\%$ on $5$, $10$ and $25$ task experiments respectively ({\color{blue_graph}\textbf{violet}} bars). We note that the gain is more significant when we compare with adding \ours to AANet with explicit regularization, corroborating the effectiveness of our 
implicit regularizer.

\customsubsection{Aligning the Final Layer Logits:}\label{sec:aligning_final_logits}
\begin{table}[b]\vspace{2pt}
\centering
\caption{Latent representations alignment is more effective than aligning logits. Subscripts show change in accuracy from baseline.}
\label{tab:logit_aligner}
\resizebox{0.47\textwidth}{!}{%
\begin{tabular}{@{}llll@{}}
\toprule
Method & 5 Tasks & 10 Tasks & 25 Tasks \\ \midrule
iCaRL~\cite{rebuffi2017icarl} & $56.97$ & $53.28$ & $50.98$ \\
iCaRL~\cite{rebuffi2017icarl} + Logit Aligner & $57.97_{\color{ForestGreen}\textbf{~+~1.00}}$ & $54.42_{\color{ForestGreen}\textbf{~+~1.14}}$ & $51.49_{\color{ForestGreen}\textbf{~+~0.51}}$ \\
iCaRL~\cite{rebuffi2017icarl} + \ours & $63.68_{\color{ForestGreen}\textbf{~+~6.71}}$ & $58.92_{\color{ForestGreen}\textbf{~+~5.64}}$ & $54.00_{\color{ForestGreen}\textbf{~+~3.02}}$ \\ \midrule
LUCIR~\cite{hou2019learning} & $64.37$ & $62.57$ & $59.91$ \\
LUCIR~\cite{hou2019learning} + Logit Aligner & $62.50_{\color{RedOrange}\textbf{~-~1.87}}$ & $61.67_{\color{RedOrange}\textbf{~-~0.9}}$ & $59.22_{\color{RedOrange}\textbf{~-~0.69}}$ \\
LUCIR~\cite{hou2019learning} + \ours & $66.06_{\color{ForestGreen}\textbf{~+~1.69}}$ & $63.50_{\color{ForestGreen}\textbf{~+~0.93}}$ & $60.30_{\color{ForestGreen}\textbf{~+~0.39}}$ \\ \midrule
AANet~\cite{liu2021adaptive} & $67.53$ & $66.25$ & $64.28$ \\
AANet~\cite{liu2021adaptive} + Logit Aligner & $66.16_{\color{RedOrange}\textbf{~-~1.37}}$ & $65.29_{\color{RedOrange}\textbf{~-~0.96}}$ & $63.81_{\color{RedOrange}\textbf{~-~0.47}}$ \\
AANet~\cite{liu2021adaptive} + \ours & $68.78_{\color{ForestGreen}\textbf{~+~1.25}}$ & $66.62_{\color{ForestGreen}\textbf{~+~0.37}}$ & $64.72_{\color{ForestGreen}\textbf{~+~0.44}}$ \\ \bottomrule
\end{tabular}%
}
\end{table}
\ours aligns the latent representations from the feature extractor
$\mathbf{z} = \mathcal{F}_{{\bm \theta}}^{\mathcal{T}_{t}}(\mathbf{x})$. An alternative would be to align the final logits $\mathcal{F}_{{\bm \phi}}^{\mathcal{T}_{t}}(\mathcal{F}_{{\bm \theta}}^{\mathcal{T}_{t}}(\mathbf{x}))$. We re-evaluate incremental CIFAR-100 experiments in this setting.
We find that latent space alignment is more effective than aligning the logit space (referred to as `+ Logit Aligner' in Tab.~\ref{tab:logit_aligner}). This is because the logits are specific to the end task while the latent representations model generalizable features across tasks. 

\customsubsection{Aligning across Different-sized Latent Spaces:}\label{sec:latent_dimention}
\ours can align latent representations of varied dimensions. Our toy experiment on MNIST uses $32$ and $512$ dimensional latent space, while CIFAR-100 experiments use a $64$ dimensional space. ImageNet and Pascal VOC experiments uses a latent space of $512$ and $2048$ dimensions each.

\customsubsection{Sensitivity to  Hyper-parameters:}\label{sec:hyper_param_search}
We alter parameters that can affect the \ours performance in  Tab.~\ref{tab:langevin},~\ref{tab:iterations} and~\ref{tab:architecture}.
The experiments are on CIFAR-100 in `iCaRL + \ours' setting. The {\colorbox{ highlight}{highlighted}} rows represent the default configuration.

\noindent\textit{Number of Langevin Steps:} In Tab.~\ref{tab:langevin}, we experiment with changing the number of Langevin steps $L_{steps}$ required to sample from EBM in Algo.~\ref{algo:align_latents}. \ours is able to align latents with very few number of steps as the energy manifold is adept in guiding the alignment of latent representations.

\noindent\textit{Number of Iterations Required:} While training the EBM using Algo.~\ref{algo:learning_ebm}, we change the number of iterations required, and report the accuracy in Tab.~\ref{tab:iterations}. At around $1000$ iterations, the EBM converges. Increasing the number of iterations further, does not lead to significant improvement.

\noindent\textit{Architecture:} We experiment with EBM models of different capacities in Tab.~\ref{tab:architecture}. We find that using a smaller architecture or 
a significantly larger architecture does not help. We see this as a desirable characteristic since we learn the energy manifold of the latent space and not the data space. 

\customsubsection{Compute and Memory: 
}\label{sec:compute_memory}
We record the compute, memory and time requirements of \ours for CIFAR-100. We use a single Nvidia Tesla K$80$ GPU for these metrics. The EBM, which is a two layer network with $64$ neurons each, has $8.385$K parameters and takes $1.057$M flops when trying to learn $64$ dimensional latent features. It takes $0.039 \pm 0.003$ secs to  sample from this EBM when aligning $64$ dimensional latents. We use $30$ Langevin iterations for sampling. Mini-batch size is $128$ for both the experiments.

\vspace{4pt}
\section{Conclusion}
\label{sec:conclusion}
\vspace{-6pt}
We demonstrate the use of energy-based models (EBMs) as a promising solution for incremental learning, by extending their natural mechanism to deal with representational shift.
This is achieved by modeling the likelihoods in the latent feature space, measuring the distributional shifts experienced across learning tasks and in-turn realigning them to optimize learning across all the tasks. 
Our proposed approach \ours, is complementary to existing methods and can be used as an add-on module without modifying their base pipeline. \ours offers consistent improvement to three prominent class-incremental classification
methodologies when evaluated across multiple settings. Further, on the harder incremental object detection task, our methodology provides significant improvement over state-of-the-art.

\vspace{-8pt}
\section*{Acknowledgements} 
\vspace{-6pt} 
\noindent \small We thank Yaoyao Liu for his prompt 
clarifications on AANET \cite{liu2021adaptive} code.
KJJ thanks TCS Research for their PhD fellowship. VNB thanks DST, Govt of India, for partially supporting this work through the IMPRINT and ICPS programs.

{\small
\bibliographystyle{ieee_fullname}
\bibliography{egbib}
}

\clearpage
\appendix
\section*{\centering Supplementary Material}
In this supplementary material, we provide additional details and experimental analysis regarding the behaviour of proposed latent alignment approach (\ours). They are:
\vspace{-3pt}
\begin{itemize}
\setlength\itemsep{-0.15em}
    \item An illustration for the adaptation process of latent representation with \ours. (Sec.~\ref{sec:implicit_importance})
    \item Effect of using \textit{mixup} for data augmentation. (Sec.~\ref{sec:data_aug})
    \item Comments on the broader societal impacts. (Sec.~\ref{sec:impact})
    \item Qualitative results on incremental detection. (Sec.~\ref{sec:qual_res})
    \item A summary of notations used in the paper. (Sec.~\ref{sec:notation})
    
\end{itemize}

\section{Recognizing Important Latents Implicitly}\label{sec:implicit_importance}
\begin{figure}[H]
\vspace{-15pt}
  \centering
  \includegraphics[width=1\columnwidth]{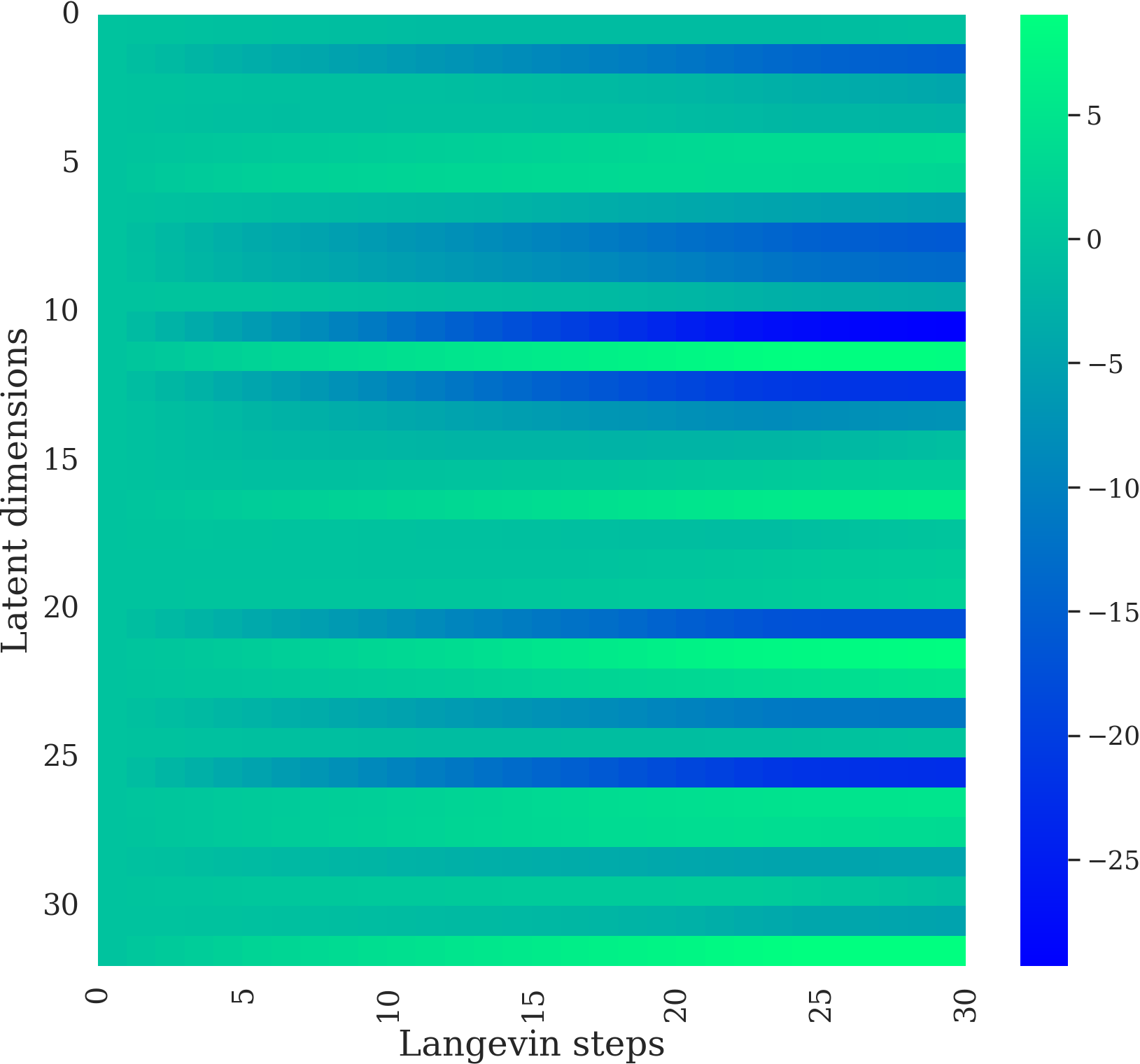}
  \caption{Each row $i$ shows how $i^{th}$ latent dimension is updated by \ours. We see that different dimensions have different degrees of change, which is implicitly decided by our energy-based model.
  }
  \label{fig:implicit_recognition}
  \vspace{-10pt}
\end{figure}
Fig.~\ref{fig:implicit_recognition} shows how each latent dimension of a $32$ dimensional latent vector (y-axis) gets adapted in each Langevin iteration (x-axis). For an initial latent representation $\mathbf{z}_0$, each column shows the difference from its aligned version from the $i^{th}$ Langevin step: $\mathbf{z}_i - \mathbf{z}_0$. We consider MNIST experiment (Sec.~3.3) for this illustration.
Our proposed latent aligner is able to implicitly identify which latent dimension is important to be preserved or modified.
This characteristic is difficult to achieve in alternate regularization methods like distillation, which gives equal weightage to each dimension. We can see that the specialization happens within a few number of iterations, similar to the results in Tab.~3. 

\section{Augmenting Data with \textit{mixup}}\label{sec:data_aug}
As detailed in Sec.~3.2, we use datapoints sampled form the current task distribution to learn the energy-based model $\mathbf{x}_{i} \sim p_{data}^{\tau_{t}}$. Here we use \textit{mixup}, an augmentation technique introduced by Zhang \etal \cite{zhang2018mixup}, where each datapoint is modified as $\mathbf{\hat{x}} = \lambda \mathbf{x}_{i} + (1 - \lambda)\mathbf{x}_{j}$, s.t. $\lambda \sim Beta(\alpha, \alpha)$, and report the results in Tab.~\ref{tab:mixup}. In these experiments with incremental CIFAR-100, we see that using \textit{mixup} does not enhance performance, even with different values of $\alpha$. This is because the EBM is a small two layer network which is not prone to overfitting, and can perform well even without this extra augmentation. 

\begin{table}[H]
\vspace{-5pt}
\centering
\caption{The performance of EBM is comparable with and without using \textit{mixup} augmentation as the EBM network is small.}
\label{tab:mixup}
\resizebox{0.37\textwidth}{!}{%
\begin{tabular}{@{}cccc@{}}
\toprule
$\alpha$         & 5 Tasks & 10 Tasks & 25 Tasks \\ \midrule
Without \textit{mixup} \cite{zhang2018mixup} & 63.68   & 58.92    & 54.00       \\
0.1           & 63.67   & 58.85    & 54.01    \\
0.3           & 63.53   & 58.81    & 53.85    \\
0.5           & 63.54   & 58.79    & 53.88    \\
1.0             & 63.44   & 58.53    & 53.83    \\ \bottomrule
\end{tabular}%
}
\vspace{-14pt}
\end{table}

\begin{figure*}[h]
\centering
\subfloat{\includegraphics[width=1.0\linewidth]{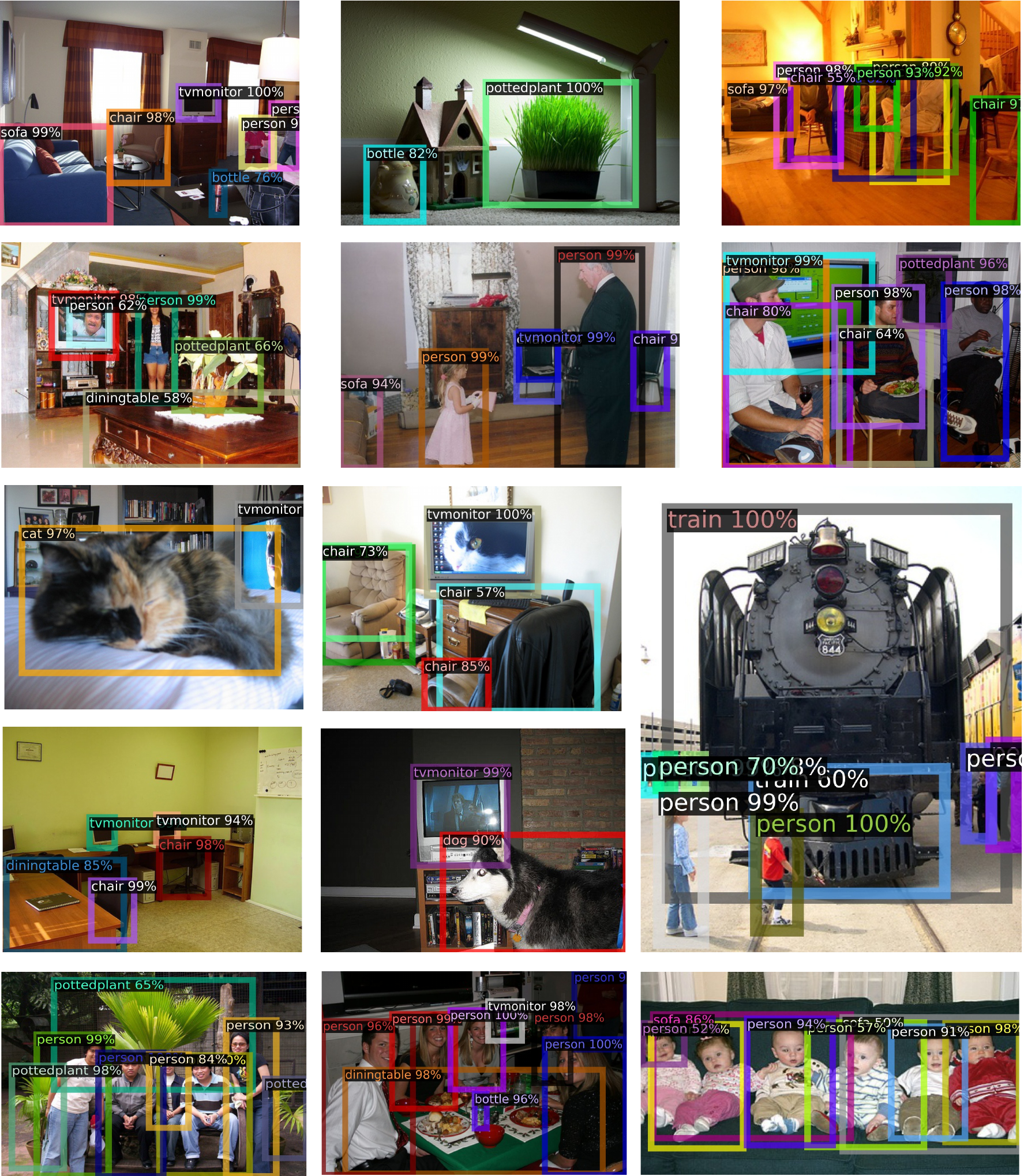}}
\caption{Qualitative results of incremental Object Detection. We consider the $10+5$ setting on Pascal VOC, where instances of  \txt{plant},  \txt{sheep},  \txt{sofa},  \txt{train} and \txt{tvmonitor} are added to a detector trained on the rest of the classes.}
\label{fig:qualitative_results}
\end{figure*}


\section{Broader Impact}\label{sec:impact}
When a model incrementally learns without forgetting, an equivalently important desiderata would be to selectively forget, in adherence to any privacy or legislative reasons. Such an unlearning can be possible by treating such instances as out-of-distribution samples, however, a dedicated treatment of the same is beyond the current scope of our work. Our current work aims to reduce the catastrophic forgetting and interference while learning continually, and to the best of our knowledge, our methodology does not have any detrimental social impacts that make us different from other research efforts geared in this direction.

\section{Qualitative Results}\label{sec:qual_res}
In Figure \ref{fig:qualitative_results}, 
we show more qualitative results for incremental Object Detection in the $15 + 5$ setting with Pascal VOC dataset \cite{everingham2010pascal}. Instances of  \txt{plant},  \txt{sheep},  \txt{sofa},  \txt{train} and \txt{tvmonitor} are added to a detector trained on the rest. The considerable improvement of \ours over the state-of-the-art-method \cite{iOD}  as shown in Tab.~2, is due to the implicit latent space regularization that \ours offers. To the best of our knowledge, \ours is the first method that adds latent space regularization to large scale incremental object detection models.

\section{Summary of Notations}\label{sec:notation}
For clarity, Tab.~\ref{tab:symbols} summarizes the main notations used in our paper along with their concise  description. 
\begin{table}[h]
\caption{To enhance readability, this table summarises the notations used in the manuscript, along with their meaning.}
\centering
\resizebox{0.48\textwidth}{!}{%
\begin{tabular}{@{}ll@{}}
\toprule
Notation & Stands for\\ \midrule
$\tau_i$    &     $i^{th}$ task    \\
  $\mathbf{x} \in \tau_{i}$    &     Image from the $i^{th}$ task    \\ $\mathcal{T}_t=\{\tau_1, \tau_2,\cdots,\tau_t\}$   &  Continuum or set of tasks seen until time $t$       \\
    $ \mathbf{x} \in  \mathcal{T}_t$   &  Image from any of the task in $\mathcal{T}_t$      \\
     $\mathcal{M}^{\mathcal{T}_t}$  &  Model trained until time $t$      \\
     $\mathcal{F}_{\bm \theta}^{\mathcal{T}_t}$ & Feature extractor of $\mathcal{M}^{\mathcal{T}_t}$
     \\
     $\mathcal{F}_{\bm \phi}^{\mathcal{T}_t}$ & Task specific part of $\mathcal{M}^{\mathcal{T}_t}$
     \\
          $\mathbf{z}^{\mathcal{T}_t}$ & Latent representation from $\mathcal{F}_{\bm \theta}^{\mathcal{T}_t}$ 
     \\
     $p_{data}^{\tau_t}$ & Data distribution of task $\tau_t$
     \\
     $(\mathbf{x}^{\tau_{t}}_i, y^{\tau_{t}}_i) \sim p_{data}^{\tau_t}$ & Samples from $p_{data}^{\tau_t}$
     \\
 \bottomrule
\end{tabular}%
}
\label{tab:symbols}
\end{table}
\end{document}